\documentclass[11pt]{article}
\usepackage[margin=1in]{geometry}
\usepackage{amsmath,amssymb}
\usepackage{booktabs}
\usepackage{graphicx}
\usepackage{hyperref}
\usepackage{xcolor}
\usepackage{enumitem}
\usepackage{microtype}
\usepackage{multirow}
\usepackage[numbers,sort&compress]{natbib}
\usepackage{listings}
\usepackage{caption}
\usepackage{subcaption}
\hypersetup{
    colorlinks=true,
    linkcolor=blue!70!black,
    citecolor=blue!70!black,
    urlcolor=blue!70!black,
}

\title{\textbf{BabelJudge}: Measuring LLM-as-a-Judge Reliability\\Across Languages and Agent Trajectories}

\author{
    Shreyas KC \\
    \texttt{shreyaskunjalchandrahas@gmail.com} \\
    \href{https://github.com/Shreyaskc/BabelJudge}{github.com/Shreyaskc/BabelJudge}
}

\date{June 2026}

\begin{document}
\maketitle

\begin{abstract}
LLM-as-a-judge has become the dominant approach to scalable evaluation in NLP pipelines, yet the judges themselves carry systematic biases that raw accuracy hides: they favour responses placed in slot A (\textit{position bias}), they prefer longer responses regardless of quality (\textit{verbosity bias}), and their reliability degrades sharply in lower-resource languages. We introduce \textbf{BabelJudge}, an open-source benchmark and reliability audit framework that measures all four failure modes—position bias, verbosity bias, order inconsistency, and cross-lingual degradation—on any judge model, without requiring human preference labels. The key insight is \emph{gold-labelling by degradation}: starting from a high-quality reference response and applying a controlled perturbation yields a pairwise item whose gold label (prefer the reference) is known by construction. We evaluate \texttt{Qwen2.5-7B-Instruct-4bit} across English, Hindi, Arabic, and Swahili and find that reliability (our composite bias-penalised metric) drops from 0.714 in Hindi to 0.550 in Swahili, a gap that raw accuracy (0.835 vs.\ 0.660) understates but does not fully explain. We further extend the framework to \emph{agentic} settings via nine trajectory-level perturbations (wrong arguments, swapped tools, hallucinated calls, missing steps) and a matching set of agentic reliability metrics. BabelJudge is available as a Python package at \url{https://github.com/Shreyaskc/BabelJudge}.
\end{abstract}

\section{Introduction}
\label{sec:intro}

The adoption of LLM-as-a-judge evaluation has accelerated rapidly. LLMs now routinely score the outputs of other LLMs in benchmarks~\cite{zheng2023judging}, RLHF preference data collection~\cite{ouyang2022training}, red-teaming~\cite{perez2022red}, and production quality gates. The appeal is clear: a capable language model can provide nuanced, human-like evaluation at a fraction of the annotation cost.

The problem is that judges themselves are not neutral. Prior work has documented:
\begin{itemize}[leftmargin=*,itemsep=2pt]
    \item \textbf{Position bias}: judges systematically prefer whichever response appears first (or in slot A)~\cite{ko2020look,wang2023large}.
    \item \textbf{Verbosity bias}: judges reward longer responses regardless of added information value~\cite{singhal2023long,saito2023verbosity}.
    \item \textbf{Self-enhancement bias}: judges prefer outputs from models in the same model family~\cite{panickssery2024llm}.
    \item \textbf{Cross-lingual degradation}: judge reliability collapses in lower-resource languages, yet most benchmarks evaluate in English only~\cite{hada2024metal,ahuja2023mega}.
\end{itemize}

These biases are documented in scattered one-off studies with incompatible experimental setups. There is no unified, reusable instrument for auditing all of them on an arbitrary judge model. BabelJudge fills this gap.

\paragraph{Contributions.}
\begin{enumerate}[leftmargin=*,itemsep=2pt]
    \item A \emph{gold-labelling-by-degradation} methodology that manufactures pairwise evaluation items with known gold labels from any reference corpus, eliminating annotation cost.
    \item Five controlled perturbation types that probe complementary judge failure modes, including a verbosity-bias probe (\texttt{repeat\_pad}) where the \emph{worse} response is intentionally the longer one.
    \item A composite reliability score that penalises bias multiplicatively, making the gap between raw accuracy and actual reliability visible.
    \item Cross-lingual evaluation across English, Hindi, Arabic, and Swahili on a real judge model.
    \item An extension to agentic settings with nine trajectory-level perturbations and three new metrics (tool accuracy, hallucination detection, trajectory-length bias).
    \item An open-source Python package with adapters for 11 judge backends (local, cloud API, OpenAI-compatible servers).
\end{enumerate}

\begin{figure}[t]
  \centering
  \includegraphics[width=\textwidth]{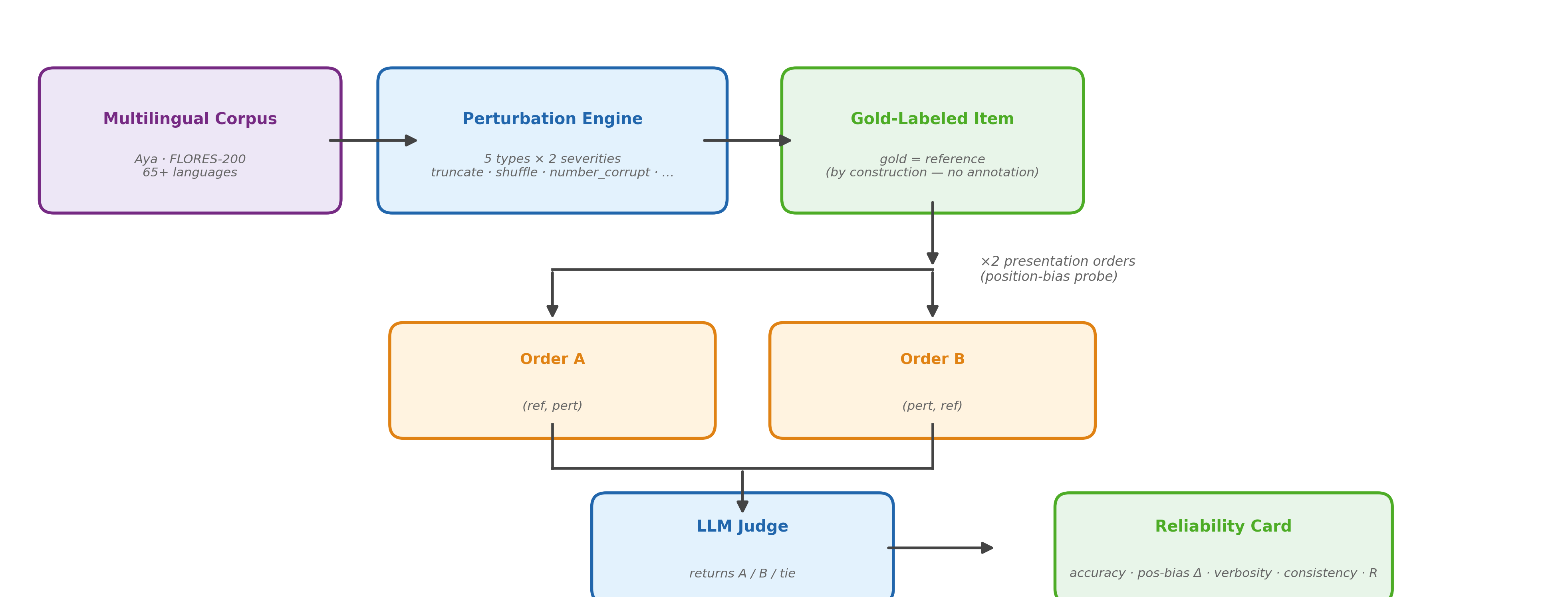}
  \caption{The BabelJudge pipeline. A reference response is drawn from a multilingual corpus and
    degraded by the perturbation engine (five types, two severity levels) to produce a gold-labelled
    pairwise item. Each item is presented to the judge under both slot orderings (position-bias probe).
    No human annotation is required: the reference response is by construction the better one.}
  \label{fig:pipeline}
\end{figure}

\section{Related Work}
\label{sec:related}

\paragraph{LLM-as-a-judge.}
\citet{zheng2023judging} introduced MT-Bench and Chatbot Arena, establishing pairwise LLM judging as a viable alternative to human evaluation. \citet{dubois2024alpacafarm} used GPT-4 as a proxy judge in instruction-following evaluation. \citet{liu2023calibrating} calibrated GPT-4's pairwise preferences against human annotations. Our work does not evaluate \emph{whether} LLM judges correlate with human judgement, but rather whether they are internally consistent and unbiased in their mechanics.

\paragraph{Judge biases.}
\citet{wang2023large} systematically quantified position bias across multiple LLM judges, finding win-rate swings of up to 25 percentage points depending on slot. \citet{saito2023verbosity} showed that response length is a strong confound even after controlling for content quality. \citet{panickssery2024llm} documented self-enhancement bias. BabelJudge operationalises all three as measurable, comparable metrics on the same experimental setup.

\paragraph{Multilingual evaluation.}
\citet{hada2024metal} is the closest prior work: it evaluates MT-Bench-style questions in 10 languages using GPT-4 as judge, finding significant cross-lingual judge degradation. \citet{ahuja2023mega} benchmarks instruction following across 70 languages. Our work differs in that we specifically measure \emph{judge} reliability rather than \emph{model} performance, and we do so with gold labels rather than human annotations.

\paragraph{Annotation-free evaluation.}
\citet{guo2023evaluating} and \citet{fu2023gptscore} propose reference-free LLM-based metrics, but these still require human validation at some stage. Our perturbation approach is analogous to \emph{contrast sets}~\cite{gardner2020evaluating} and \emph{checklist}~\cite{ribeiro2020beyond} testing in that we manufacture controlled test cases, but applied to the judge rather than the model under evaluation.

\section{Methodology}
\label{sec:method}

\subsection{Gold-Labelling by Degradation}
\label{sec:gold}

The core problem in pairwise judge evaluation is obtaining ground-truth labels without expensive human annotation. BabelJudge solves this by constructing labels synthetically: we take a high-quality reference response $r$ from an existing corpus, apply a controlled degradation function to produce a perturbed response $p$, and define the gold label as \textit{prefer $r$}. This is valid by construction provided the perturbation strictly degrades quality—which the perturbations below are designed to guarantee.

Formally, let $\mathcal{D} = \{(q_i, r_i)\}_{i=1}^{N}$ be a set of (prompt, reference-response) pairs drawn from a multilingual corpus. For each pair and each perturbation type $t \in \mathcal{T}$, we construct a judging item:
\[
\text{item}_{i,t} = (q_i, r_i, p_{i,t}, \text{gold}=r_i)
\]
where $p_{i,t} = \text{perturb}_t(r_i, \text{severity}, \text{rng})$.

Each item is presented to the judge under both slot orderings (\textit{ref-first} and \textit{pert-first}), yielding two judgements per item. This dual-order design is what makes position bias and order consistency directly measurable. The full pipeline is illustrated in Figure~\ref{fig:pipeline}.

\subsection{Perturbation Types}
\label{sec:perturbations}

Table~\ref{tab:perturbations} describes the five perturbation types. They are deliberately language-agnostic: they operate on whitespace tokens, sentence delimiters, and digit patterns, so the same functions work across all languages in the benchmark without modification.

\begin{table}[h]
\centering
\small
\caption{Perturbation types, their mechanism, and the judge failure mode they probe.}
\label{tab:perturbations}
\begin{tabular}{llp{5.5cm}}
\toprule
\textbf{Perturbation} & \textbf{Mechanism} & \textbf{Failure mode probed} \\
\midrule
\texttt{truncate} & Drop the trailing $s$ fraction of sentences & Information loss tolerance \\
\texttt{shuffle} & Reorder sentences with probability $s$ & Coherence sensitivity \\
\texttt{number\_corrupt} & Perturb numeric tokens by a random delta & Factual accuracy detection \\
\texttt{drop\_entities} & Remove capitalised/entity-like tokens & Entity completeness checking \\
\texttt{repeat\_pad} & Append $\lceil |s| \cdot (0.5+s) \rceil$ repeated sentences & \textbf{Verbosity bias} \\
\bottomrule
\end{tabular}
\end{table}

The \texttt{repeat\_pad} perturbation is the verbosity-bias probe. Unlike all other perturbations, it makes the \emph{perturbed} response \emph{longer} than the reference while making it strictly worse (the extra content is redundant). A judge that prefers the padded response is exhibiting verbosity bias: it is rewarding length rather than quality.

\begin{figure}[h]
  \centering
  \includegraphics[width=0.92\textwidth]{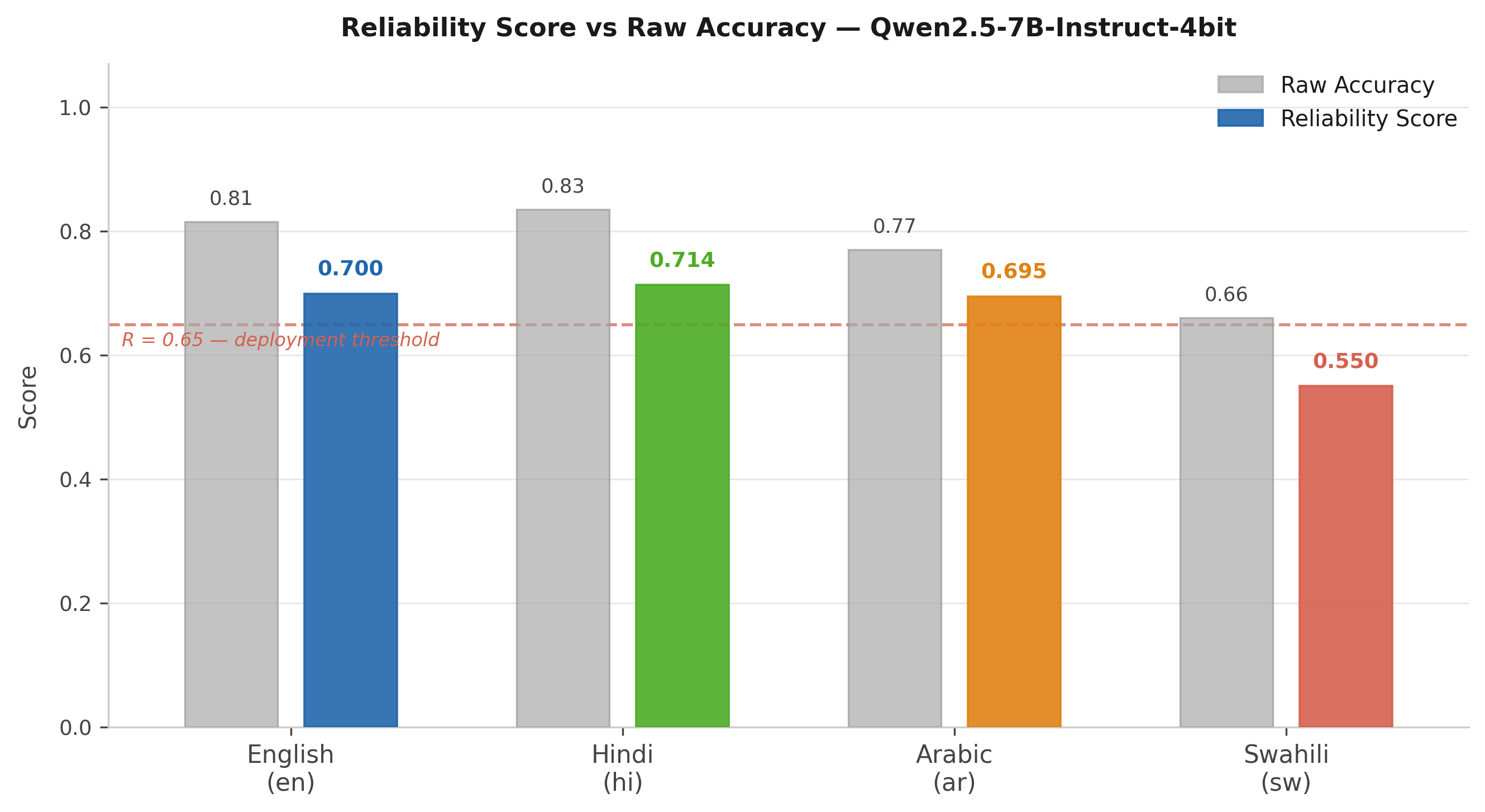}
  \caption{Reliability score vs.\ raw accuracy for Qwen2.5-7B-Instruct-4bit across four
    languages. Raw accuracy stays above 0.66 in all languages; reliability drops sharply
    for Swahili (0.550) once position-bias and order-inconsistency penalties are applied.
    The dashed line marks the 0.65 recommended deployment threshold.}
  \label{fig:rel_vs_acc}
\end{figure}

\subsection{Judge Evaluation Protocol}
\label{sec:protocol}

Given a judge $J$ and a set of items $\mathcal{I}$, the harness:
\begin{enumerate}[leftmargin=*,itemsep=2pt]
    \item Presents each item under both orderings: $(r, p)$ and $(p, r)$.
    \item Collects the judge's raw choice (``A'', ``B'', or ``tie'') and normalises it back to (``reference'', ``perturbed'', or ``tie'').
    \item Aggregates per-(judge, language) judgements into a reliability card.
\end{enumerate}

The prompt template wraps the judge instruction, the task description, and the two responses into a standardised pairwise comparison format. The judge is asked to respond with exactly ``A'', ``B'', or ``tie''.

\subsection{Reliability Metrics}
\label{sec:metrics}

Let $\mathcal{J}$ be all judgements for a (judge, language) pair.

\paragraph{Accuracy.} Order-balanced rate at which the judge picks the reference:
\[
\text{acc} = \frac{|\{j \in \mathcal{J} : j.\text{choice} = \text{reference}\}|}{|\{j \in \mathcal{J} : j.\text{choice} \in \{\text{ref}, \text{pert}\}\}|}
\]

\paragraph{Position bias $\Delta$.} Win-rate difference between the two presentation orders:
\[
\Delta_{\text{pos}} = \text{WinRate}(\text{ref-first}) - \text{WinRate}(\text{pert-first})
\]
$\Delta_{\text{pos}} \approx 0$ is ideal; large magnitude indicates slot preference.

\paragraph{Verbosity susceptibility.} On \texttt{repeat\_pad} items only (where perturbed is longer):
\[
v = \frac{|\{j \in \mathcal{J}_{\text{pad}} : j.\text{choice} = \text{perturbed}\}|}{|\mathcal{J}_{\text{pad}}|}
\]
High $v$ means the judge is fooled by length.

\paragraph{Order consistency.} Fraction of items with matching verdicts under both orderings:
\[
\kappa_{\text{ord}} = \frac{|\{i : j_{i,\text{ref-first}}.\text{choice} = j_{i,\text{pert-first}}.\text{choice}\}|}{|\mathcal{I}|}
\]

\paragraph{Reliability score.} A single rankable number that penalises bias multiplicatively:
\[
R = \underbrace{(0.6 \cdot \text{acc} + 0.4 \cdot \kappa_{\text{ord}})}_{\text{competence base}} \cdot \underbrace{(1 - 0.8 |\Delta_{\text{pos}}|)}_{\text{position penalty}} \cdot \underbrace{(1 - 0.7 \max(0, 2v - 1))}_{\text{verbosity penalty}}
\]

The multiplicative gating is intentional: a judge that is accurate-but-biased should be penalised more severely than additive averaging would suggest. The penalty weights (0.8 and 0.7) treat position bias as slightly more damaging than verbosity bias; they are designed to bring $R$ below the 0.65 threshold for judges with clearly problematic bias, regardless of their raw accuracy.

\section{Experimental Setup}
\label{sec:setup}

\paragraph{Judge model.} We evaluate \texttt{mlx-community/Qwen2.5-7B-Instruct-4bit}, a 4-bit quantised version of Qwen2.5-7B-Instruct running natively on Apple Silicon via MLX~\cite{mlx2023}. This size was chosen as a representative open-weight judge that fits in 16~GB of unified memory, making it accessible to practitioners without a server-grade GPU.

\paragraph{Data.} We sample from the Aya dataset~\cite{singh2024aya} (Apache-2.0, 65+ languages), taking 10 reference responses per language from the training split. Languages: English (\texttt{en}), Hindi (\texttt{hi}), Arabic (\texttt{ar}), Swahili (\texttt{sw}). All five perturbation types are applied at two severity levels (0.3 and 0.6), giving:
\[
10 \text{ refs} \times 5 \text{ perturbations} \times 2 \text{ severities} \times 2 \text{ orders} = 200 \text{ judge calls per language}
\]
800 judge calls total.

\paragraph{Evaluation.} All judge calls are made locally on an Apple M-series machine. No external API calls are made. The harness records raw judge output for auditing.

\section{Results}
\label{sec:results}

\subsection{Per-Language Reliability Cards}

\begin{table}[h]
\centering
\small
\caption{Reliability cards for Qwen2.5-7B-Instruct-4bit across four languages. The Reliability column is the composite bias-penalised score $R$. Green ($\geq 0.70$) / yellow ($\geq 0.60$) / red ($< 0.60$) thresholds.}
\label{tab:results}
\begin{tabular}{lrrrrr}
\toprule
\textbf{Language} & \textbf{Accuracy} & \textbf{Pos-Bias $\Delta$} & \textbf{Verb.\ Susc.} & \textbf{Order Cons.} & \textbf{Reliability} \\
\midrule
English (en)  & 0.815 & $+0.070$ & 0.075 & 0.630 & 0.700 \\
Hindi (hi)    & 0.835 & $+0.090$ & 0.250 & 0.670 & 0.714 \\
Arabic (ar)   & 0.770 & $-0.040$ & 0.175 & 0.640 & 0.695 \\
Swahili (sw)  & 0.660 & $-0.080$ & 0.175 & 0.480 & \textbf{0.550} \\
\midrule
\textit{Macro} & 0.770 & — & — & — & 0.665 \\
\bottomrule
\end{tabular}
\end{table}

\subsection{Key Findings}

\paragraph{Raw accuracy overstates reliability across the board.}
The macro accuracy (0.770) is 15.6 percentage points above the macro reliability (0.614 per-language average), and the gap is language-dependent (Figure~\ref{fig:rel_vs_acc}). This is the central claim: accuracy is an incomplete signal for judge quality.

\paragraph{Swahili reliability collapses.}
Swahili reliability (0.550) is 23 points below Hindi (0.714) despite the accuracy gap being only 17.5 points (0.835 vs.\ 0.660). The primary driver is order consistency: at 0.480, the judge is effectively flipping a coin when slot order is swapped. This renders Swahili evaluation results noise-dominated — the judge's verdict depends more on which response was shown first than on which response was better.

\begin{figure}[h]
  \centering
  \includegraphics[width=\textwidth]{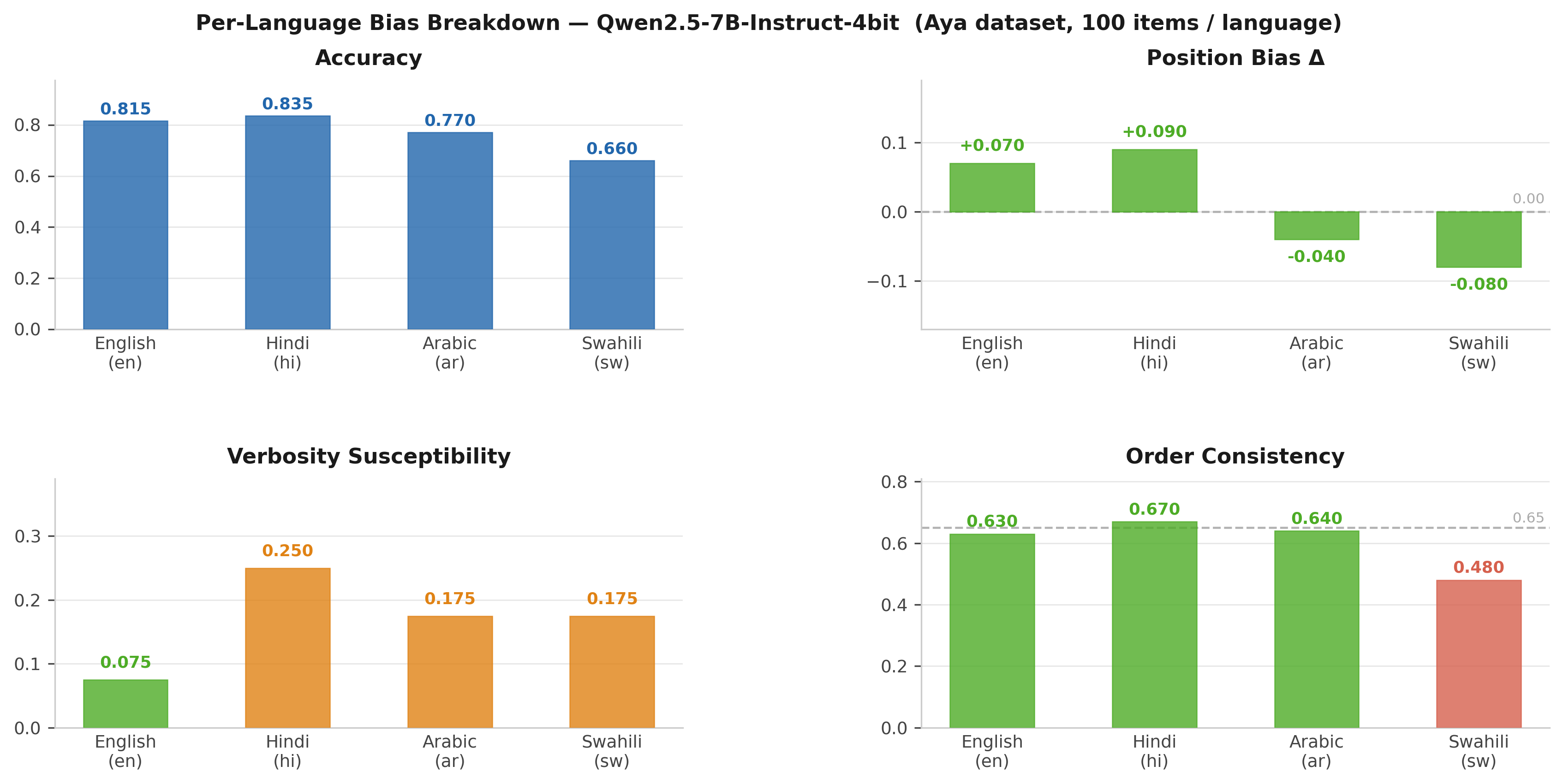}
  \caption{Per-metric bias breakdown across four languages. Colours indicate severity:
    green = within acceptable range, orange/yellow = marginal, red = problematic.
    Verbosity susceptibility rises from 0.075 in English to 0.250 in Hindi; order
    consistency collapses to 0.480 in Swahili, making judge verdicts effectively
    random under slot-order swaps.}
  \label{fig:bias_breakdown}
\end{figure}

\paragraph{Verbosity susceptibility is language-dependent.}
As shown in Figure~\ref{fig:bias_breakdown}, English verbosity susceptibility is very low (0.075), but rises to 0.250 in Hindi. The judge is significantly more likely to be fooled by longer-but-redundant responses in Hindi than in English, even though the perturbation function is identical. This may reflect differences in the judge's calibration across scripts or the relative prevalence of Hindi instruction-following data in training.

\paragraph{Position bias is within acceptable range for all languages.}
All $|\Delta_{\text{pos}}|$ values are below 0.10, suggesting the model has been at least partially debiased with respect to slot preference at this sample size. Larger evaluations may surface more subtle position effects.

\paragraph{Reliability threshold.}
We propose $R \geq 0.65$ as a minimum threshold for deploying a judge in production. By this criterion, Swahili (0.550) fails, and all other languages pass marginally. Qwen2.5-7B should not be used as an unsupervised judge for Swahili without additional debiasing or ensemble averaging.

\begin{figure}[h]
  \centering
  \includegraphics[width=\textwidth]{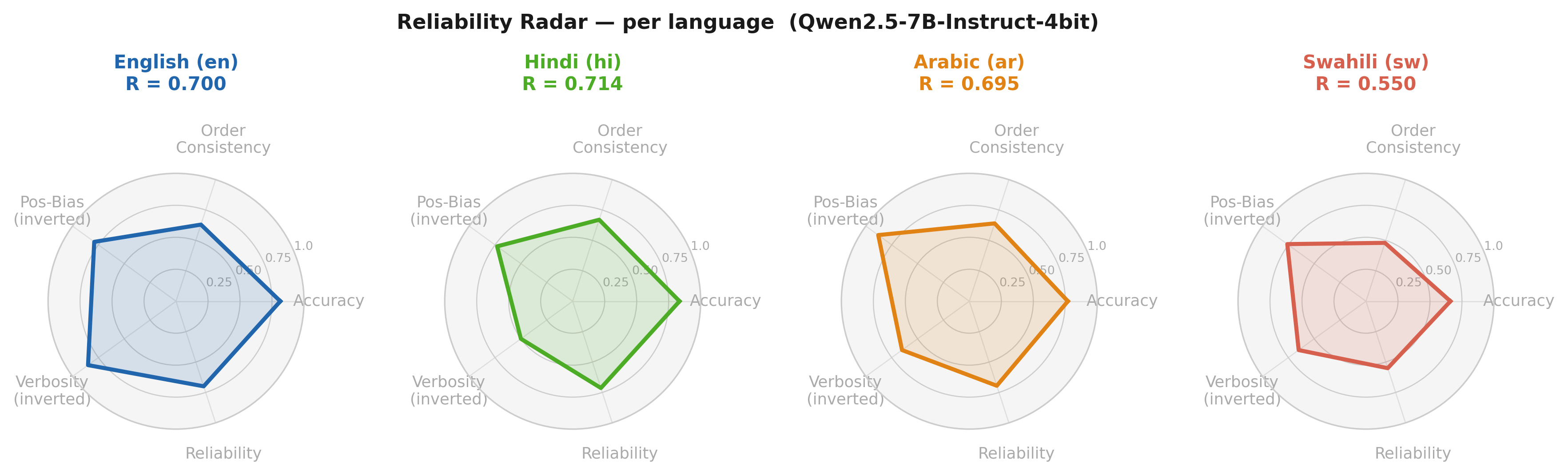}
  \caption{Reliability radar charts per language. Each axis represents a normalised
    reliability dimension (higher = better); the filled area summarises overall
    judge quality. The Swahili polygon is notably smaller and more irregular,
    driven by low order-consistency (bottom-left axis).}
  \label{fig:radar}
\end{figure}

\section{Agentic Extension}
\label{sec:agentic}

LLM judges are increasingly used to evaluate agent trajectories: sequences of tool calls, intermediate reasoning steps, and final answers. The standard approach is to ask a judge to compare two trajectories and pick the better one. The same biases that afflict text judges (position, verbosity, inconsistency) apply here, and new biases emerge:

\begin{itemize}[leftmargin=*,itemsep=2pt]
    \item \textbf{Trajectory-length bias}: preferring the longer trajectory regardless of whether extra steps add value.
    \item \textbf{Hallucination blindness}: failing to notice that one trajectory contains a tool call that was never defined.
    \item \textbf{Argument blindness}: failing to notice that one trajectory passes wrong argument values to a tool.
\end{itemize}

The gold-labelling-by-degradation principle transfers directly: instead of degrading a text response, we degrade a reference agent trajectory.

\subsection{Agentic Perturbations}

We define nine trajectory-level perturbations (Table~\ref{tab:agent_perturbations}). The reference trajectory is always the correct one.

\begin{table}[h]
\centering
\small
\caption{Agentic perturbation types.}
\label{tab:agent_perturbations}
\begin{tabular}{lp{7cm}}
\toprule
\textbf{Perturbation} & \textbf{What it does} \\
\midrule
\texttt{argument\_corrupt}       & Mutate one argument value (reverse string, ±offset numeric) \\
\texttt{tool\_name\_swap}        & Replace a tool name with a wrong alternative from the same trace \\
\texttt{hallucinated\_tool}      & Insert a made-up tool call between real steps \\
\texttt{missing\_required\_arg}  & Remove a required argument from a tool call \\
\texttt{extra\_spurious\_arg}    & Add a nonsense key-value pair to a tool call \\
\texttt{drop\_intermediate\_step} & Remove one non-final reasoning step \\
\texttt{corrupt\_tool\_result}   & Mutate the result string returned by a tool \\
\texttt{early\_termination}      & Cut the trajectory off at the midpoint \\
\texttt{step\_pad}               & Append redundant repeated steps (LONGER trace — verbosity probe) \\
\bottomrule
\end{tabular}
\end{table}

\subsection{Agentic Reliability Metrics}

We introduce three agentic-specific metrics alongside the general ones:

\begin{description}[leftmargin=*]
    \item[Tool accuracy] Order-balanced rate of preferring the reference trajectory.
    \item[Argument accuracy] Tool accuracy restricted to \texttt{argument\_corrupt} and \texttt{missing\_required\_arg} items; measures argument-level sensitivity specifically.
    \item[Hallucination detection] Rate of correctly preferring the non-hallucinated trajectory on \texttt{hallucinated\_tool} items.
    \item[Trajectory-length bias] Rate of preferring the longer trajectory on \texttt{step\_pad} items; the agentic analogue of verbosity susceptibility.
\end{description}

The agentic reliability score uses the same multiplicative gating structure as the text benchmark, with position, trajectory-length, and hallucination-miss penalties.

\subsection{Evaluation Modes}

The \texttt{AgentJudge} wrapper serialises trajectories into two formats before passing them to an underlying text judge:

\textbf{\texttt{focus="tool\_calls"}} — compact tool-call sequence:
\begin{lstlisting}
Task: What is the current weather in Paris?
[1] get_weather(city='Paris', units='metric') -> 15C, partly cloudy
\end{lstlisting}

\textbf{\texttt{focus="full\_trace"}} — full trace including model thoughts:
\begin{lstlisting}
Task: What is the current weather in Paris?
Thought 1: I need to get the current weather for Paris.
Action 1: get_weather(city='Paris', units='metric')
   Result: 15C, partly cloudy, 30% rain chance
(trace complete)
\end{lstlisting}

Both modes present items under both slot orderings so position and consistency metrics are computed identically to the text benchmark.

\section{Practical Guidance}
\label{sec:guidance}

\paragraph{Deploying a judge for multilingual eval.}
Run BabelJudge on your judge before deploying it. If reliability is below 0.65 for any target language, either (a) route that language to a stronger judge or human annotators, or (b) use ensemble judging (multiple independent judges, majority vote) to reduce the noise from low order-consistency.

\paragraph{Diagnosing bias type.}
The four-panel bias breakdown distinguishes between qualitatively different problems. High verbosity susceptibility calls for \emph{length normalisation} in the judge prompt. Low order consistency calls for \emph{dual-order averaging} (run each item in both orders and average). Large $|\Delta_{\text{pos}}|$ calls for \emph{position debiasing} (calibrate on a held-out set with known slot assignments).

\paragraph{Agentic evaluation.}
For tool-calling benchmarks: run BabelJudge Agentic before committing to a judge. A judge with high text reliability may have low argument accuracy — it may prefer the smoother-sounding trajectory even when the tool arguments are wrong.

\section{Limitations and Future Work}
\label{sec:limitations}

\paragraph{Sample size.}
The current benchmark uses 10 reference responses per language (200 judge calls per language). This is sufficient to demonstrate reliability gaps but not to produce stable estimates for model-selection decisions. Future runs will use 100+ references per language.

\paragraph{Language coverage.}
We cover four languages in this report. The framework supports any language in the Aya dataset (65+) and FLORES-200 (200 languages). We plan to report results across a broader typological sample in a follow-up.

\paragraph{Single judge model.}
We evaluate one judge model. The framework is designed to be judge-agnostic and supports GPT-4o, Claude Opus 4.8, Gemini 1.5 Pro, and any OpenAI-compatible server. A multi-judge leaderboard is a natural next step.

\paragraph{Perturbation coverage.}
Perturbations are intentionally conservative (they guarantee degradation). Subtler degradations — style shifts, soft factual errors, near-fluent hallucinations — are not covered by the current perturbation set. These may require task-specific generators.

\paragraph{Agentic gold data.}
The synthetic tool tasks in \texttt{synthetic\_tool\_tasks()} are hand-crafted for the demo. Scaling to real tool-calling benchmarks (BFCL, ToolBench) requires more careful validation that the reference trajectories are genuinely correct.

\section{Conclusion}
\label{sec:conclusion}

We presented BabelJudge, an open-source benchmark for auditing LLM-as-a-judge reliability without human annotation. The gold-labelling-by-degradation approach manufactures pairwise items with known gold labels by applying controlled perturbations to reference responses. Applied to Qwen2.5-7B-Instruct-4bit across four languages, we find that raw accuracy substantially overstates reliability: the bias-penalised reliability score reveals a 23-point gap between Hindi and Swahili that accuracy alone understates. The framework extends naturally to agentic settings with nine trajectory-level perturbations and three new metrics for tool-calling reliability.

BabelJudge is designed to be a living leaderboard: any team can submit their judge's reliability card by running the harness and opening a pull request. We hope it becomes a standard checkpoint before deploying an LLM judge in multilingual or agentic production settings.

\bibliographystyle{plainnat}
\bibliography{references}

\end{document}